\documentclass[11pt,a4paper]{article}

\usepackage{amsmath}
\usepackage{graphicx}
\usepackage{multirow}
\usepackage{rotating}
\usepackage{subfigure}

\usepackage{bm}

\DeclareMathOperator*{\argmax}{\arg\!\max}
\DeclareGraphicsExtensions{.png,.jpg}

\usepackage{times}

\usepackage{algorithm}
\usepackage{algorithmic}
\usepackage{authblk}

\usepackage{hyperref}

\newcommand{\overbar}[1]{\mkern 1.5mu\overline{\mkern-1.5mu#1\mkern-1.5mu}\mkern 1.5mu}

\begin{document}
%



\title{DeepSat -- A Learning framework for Satellite Imagery}
%
%
%
%
%

%

\author[1]{Saikat Basu}
\author[2]{Sangram Ganguly}
\author[1]{Supratik Mukhopadhyay}
\author[1]{Robert DiBiano}
\author[1]{Manohar Karki}
\author[3]{Ramakrishna Nemani}
\affil[1]{Department of Computer Science, Louisiana State University}
\affil[2]{Bay Area Environmental Research Institute/ NASA Ames Research Center}
\affil[3]{NASA Advanced Supercomputing Division/ NASA Ames Research Center}

\renewcommand\Authands{ and }

\maketitle
\begin{abstract}
Satellite image classification is a challenging problem that lies at the crossroads of remote sensing, computer vision, and machine learning. Due to the high variability inherent in satellite data, most of the current object classification approaches are not suitable for handling satellite datasets. The progress of satellite image analytics has also been inhibited by the lack of a single labeled high-resolution dataset with multiple class labels. The contributions of this paper are twofold -- (1) first, we present two new satellite datasets called SAT-4 and SAT-6, and (2) then, we propose a classification framework that extracts features from an input image, normalizes them and feeds the normalized feature vectors to a Deep Belief Network for classification. On the SAT-4 dataset, our best network produces a classification accuracy of 97.95\% and outperforms three state-of-the-art object recognition algorithms, namely - Deep Belief Networks, Convolutional Neural Networks and Stacked Denoising Autoencoders by $\sim$11\%. On SAT-6, it produces a classification accuracy of 93.9\% and outperforms the other algorithms by $\sim$15\%. Comparative studies with a Random Forest classifier show the advantage of an unsupervised learning approach over traditional supervised learning techniques. A statistical analysis based on Distribution Separability Criterion and Intrinsic Dimensionality Estimation substantiates the effectiveness of our approach in learning better representations for satellite imagery. 
\end{abstract}

\vspace{20 mm}

%


\section{Introduction}
\emph{Deep Learning} has gained popularity over the last decade due to its ability to learn data representations in an unsupervised manner and generalize to unseen data samples using hierarchical representations. The most recent and best-known \emph{Deep learning model} is the \emph{Deep Belief Network} ~\cite{Hinton06afast}. Over the last decade, numerous breakthroughs have been made in the field of Deep Learning; a notable one being ~\cite{QuocCCDN12}, where a locally connected sparse autoencoder was used to detect objects in the ImageNet dataset~\cite{Deng09imagenet} producing state-of-the-art results. In ~\cite{MohamedDH12}, Deep Belief Networks have been used for modeling acoustic signals and have been shown to outperform traditional approaches using Gaussian Mixture Models for Automatic Speech Recognition (ASR). They have also been found useful in hybrid learning models for noisy handwritten digit classification ~\cite{basuesann2015}. Another closely related approach, which has gained much traction over the last decade, is the Convolutional Neural Network~\cite{Lecun98gradient-basedlearning}. This has been shown to outperform Deep Belief Network in classical object recognition tasks like MNIST~\cite{mnist}, and CIFAR~\cite{Krizhevsky09learningmultiple}.

A related and equally hard problem is Satellite\footnote{Note that we use the terms satellite and airborne interchangeably in this paper because the extracted features and  learning algorithms are generic enough to handle both satellite and airborne datasets.} image classification. It involves terabytes of data and significant variations due to conditions in data acquisition, pre-processing and filtering. Traditional supervised learning methods like Random Forests~\cite{Breiman:2001} do not generalize well for such a large-scale learning problem. A novel classification algorithm for detecting roads in Aerial imagery using Deep Neural Networks was proposed in ~\cite{MnihHinton2010}. The problem of detecting various land cover classes in general is a difficult problem considering the significantly higher intra-class variability in land cover types such as trees, grasslands, barren lands, water bodies, etc. as compared to that of roads. Also, in ~\cite{MnihHinton2010}, the authors used a window of size 64$\times$64 to derive contextual information. For our general classification problem, a 64$\times$64 window is too big a context covering a total area of 64m$\times$64m. A tree canopy, or a grassy patch can typically be much smaller than this area and hence we are constrained to use a contextual window having a maximum dimension of 28m$\times$28m.

Traditional supervised learning approaches require carefully selected handcrafted features and substantial amounts of labeled data. On the other hand, purely unsupervised approaches are not able to learn the higher order dependencies inherent in the land cover classification problem. So, we propose a combination of handcrafted features that were first used in ~\cite{haralick1973} and an unsupervised learning framework using Deep Belief Network ~\cite{Hinton06afast} that can learn data representations from large amounts of unlabeled data.

There has been limited research in the field of satellite image classification due to a dearth of labeled satellite image datasets. The most well known labeled satellite dataset is the NLCD 2006 ~\cite{wickham2013}, which covers the entire globe and provide a spatial resolution of 30m. However, at this resolution, it becomes extremely difficult to distinguish between various landcover types. A high-resolution dataset acquired at a spatial resolution of 1.2m was used in ~\cite{MnihHinton2010}. However, the total area covered by the datasets namely URBAN1 and URBAN2 was ${\sim}600$ square kilometers, which included both training and testing datasets. The labeling was also available only for roads. Satellite/airborne image classification at a spatial resolution of 1-m was addressed in \cite{basu2015}. However, they performed tree-cover delineation by training a binary classifier based on Feedforward Backpropagation Neural Networks. 

The main contributions of our work are twofold -- (1) We first present two labeled datasets of airborne images -- SAT-4 and SAT-6 covering a total area of ${\sim}800$ square kilometers, which can be used to further the research and investigate the use of various learning models for airborne image classification. Both SAT-4 and SAT-6 are sampled from a much larger dataset~\cite{naip}, which covers the whole of continental United States and can be used to create labeled landcover maps, which can then be used for various applications such as measuring ground carbon content or estimating total area of rooftops for solar power generation. 

(2) Next, we present a framework for the classification of satellite/airborne imagery that a) extracts features from the image, b) normalizes the features, and c) feeds the normalized feature vectors to a Deep Belief Network for classification. On the SAT-4 dataset, our framework outperforms three state-of-the-art object recognition algorithms - Deep Belief Networks, Convolutional Neural Networks and Stacked Denoising Autoencoders by $\sim$11\% and produces an accuracy of 97.95\%. On SAT-6, it produces an accuracy of 93.9\% and outperforms the other algorithms by $\sim$15\%. We also present a statistical analysis based on Distribution Separability Criterion and Intrinsic Dimensionality Estimation to justify the effectiveness of our feature extraction approach to obtain better representations for satellite data.      

\section[Dataset]{Dataset\footnote{The SAT-4 and SAT-6 datasets are available at the web link \cite{datasets}}}
\begin{figure*}
  \centering
    \includegraphics[width=0.7\textwidth]{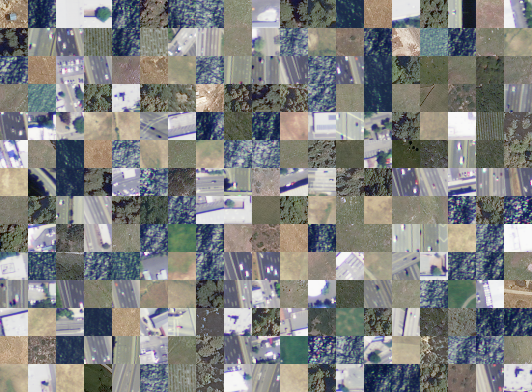}
  \caption{Sample images from the SAT-6 dataset} \label{Fig:dataset}
\end{figure*}
Images were extracted from the National Agriculture Imagery Program (NAIP \cite{naip}) dataset. The NAIP dataset consists of a total of 330,000 scenes spanning the whole of the Continental United States (CONUS). We used the uncompressed digital Ortho quarter quad tiles (DOQQs) which are GeoTIFF images and the area corresponds to the United States Geological Survey (USGS) topographic quadrangles. The average image tiles are $\sim$6000 pixels in width and $\sim$7000 pixels in height, measuring around 200 megabytes each. The entire NAIP dataset for CONUS is $\sim$65 terabytes. The imagery is acquired at a ground sample distance (GSD) of 1 meter. The horizontal accuracy lies within 6 meters of ground control points identifiable from the acquired imagery \cite{modis}. The images consist of 4 bands -- red, green, blue and Near Infrared (NIR). In order to maintain the high variance inherent in the entire NAIP dataset, we sample image patches from a multitude of scenes (a total of 1500 image tiles) covering different landscapes like rural areas, urban areas, densely forested, mountainous terrain, small to large water bodies, agricultural areas, etc. covering the whole state of California. An image labeling tool developed as part of this study was used to manually label uniform image patches belonging to a particular landcover class. Once labeled, $28{\times}28$ non-overlapping sliding window blocks were extracted from the uniform image patch and saved to the dataset with the corresponding label. We chose $28{\times}28$ as the window size to maintain a significantly bigger context as pointed by \cite{MnihHinton2010}, and at the same time not to make it as big as to drop the relative statistical properties of the target class conditional distributions within the contextual window. Care was taken to avoid interclass overlaps within a selected and labeled image patch. Sample images from the dataset are shown in Figure \ref{Fig:dataset}.

\subsection{SAT-4}
SAT-4 consists of a total of 500,000 image patches covering four broad land cover classes. These include -- barren land, trees, grassland and a class that consists of all land cover classes other than the above three. 400,000 patches (comprising of four-fifths of the total dataset) were chosen for training and the remaining 100,000 (one-fifths) were chosen as the testing dataset. We ensured that the training and test datasets belong to disjoint set of image tiles. Each image patch is size normalized to $28{\times}28$ pixels. Once generated, both the training and testing datasets were randomized using a pseudo-random number generator.

\subsection{SAT-6}
SAT-6 consists of a total of 405,000 image patches each of size $28{\times}28$ and covering 6 landcover classes - barren land, trees, grassland, roads, buildings and water bodies. 324,000 images (comprising of four-fifths of the total dataset) were chosen as the training dataset and 81,000 (one fifths) were chosen as the testing dataset. Similar to SAT-4, the training and test sets were selected from disjoint NAIP tiles. Once generated, the images in the dataset were randomized in the same way as that for SAT-4. The specifications for the various landcover classes of SAT-4 and SAT-6 were adopted from those used in the National Land Cover Data (NLCD) algorithm \cite{nlcd_desc}.

\section{Investigation of various \\
Deep Learning Models}

\subsection{Deep Belief Network} \label{sec:DBN}
Deep Belief Network (DBN) consists of multiple layers of stochastic, latent variables trained using an unsupervised learning algorithm followed by a supervised learning phase using feedforward backpropagation Neural Networks.  In the unsupervised pre-training stage, each layer is trained using a Restricted Boltzmann Machine (RBM). Unsupervised pre-training is an important step in solving a classification problem with terabytes of data and high variability.  A DBN is a graphical model \cite{Koller2009} where neurons of the hidden layer are conditionally independent of one another for a particular configuration of the visible layer and vice versa. A DBN can be trained layer-wise by iteratively maximizing the conditional probability of the input vectors or visible vectors given the hidden vectors and a particular set of layer weights. As shown in \cite{Hinton06afast}, this layer-wise training can help in improving the variational lower bound on the probability of the input training data, which in turn leads to an improvement of the overall generative model. 

We first provide a formal introduction to the Restricted Boltzmann Machine.  
The RBM can be denoted by the energy function:
\begin{equation}
E(v,h) = -\sum_{i} a_i v_i - \sum_{j} b_j h_j - \sum_{i} \sum_{j} h_j w_{i,j} v_i 
\end{equation}

where, the RBM consists of a matrix of layer weights $W=(w_{i,j})$ between the hidden units $h_j$ and the visible units $v_i$. The $a_i$ and $b_j$ are the bias weights for the visible units and the hidden units respectively.
The RBM takes the structure of a bipartite graph and hence it only has inter-layer connections between the hidden or visible layer neurons but no intra-layer connections within the hidden or visible layers. So, the activations of the visible unit neurons are mutually independent for a given set of hidden unit activations and vice versa \cite{carreiraperpinan2005contrastive}.  Hence, by setting either $h$ or $v$ constant, we can compute the conditional distribution of the other as follows:

\begin{equation}
P(h_j=1|v) = \sigma(b_j + \sum_{i=1}^{m} w_{i,j} v_{i})
\end{equation}

\begin{equation}
P(v_i=1|h) = \sigma(a_i + \sum_{j=1}^{n} w_{i,j} h_{j})
\end{equation}

where, $\sigma$ denotes the log sigmoid function:

\begin{equation}
f(x) = \frac{1}{1+e^{-x}}
\end{equation}

The training algorithm maximizes the expected log probability assigned to the training dataset $V$. So if the training dataset $V$ consists of the visible vectors $v$, then the objective function is as follows:

\begin{equation}
\argmax_{W} E\Big[\sum_{v \in V} \log{P(v)} \Big]
\end{equation}

A RBM is trained using a \emph{Contrastive Divergence} algorithm \cite{carreiraperpinan2005contrastive}. Once trained, the DBN can be used to initialize the weights of the Neural Network for the supervised learning phase \cite{Bengio2009}. 

Next, we investigate the classification accuracy of various architectures of DBN on both SAT-4 and SAT-6 datasets.

\subsubsection{DBN Results on SAT-4 \& SAT-6}
To investigate the performance of the DBN, we experiment with both \emph{big} and \emph{deep} neural architectures. This is done by varying the number of neurons per layer as well as the total number of layers in the network. Our objective is to investigate whether the more complex features learned in the deeper layers of the DBN are able to provide the network with the discriminative power required to handle higher-order texture features typical of satellite imagery data. The results from the DBN for various network architectures for SAT-4 and SAT-6 are enumerated in Table \ref{table:DBN_accuracy_SAT_4_and_SAT_6}. Each network was trained for a maximum of 500 epochs and the network state with the lowest validation error was used for testing. Regularization is done using $L_2$ norm-regularization. It can be seen from the table that for both SAT-4 and SAT-6, the classifier accuracy initially improves and then falls as more neurons or layers are added to the network.

\begin{table}[h]
\centering
\begin{tabular}{ | c | c | c | c | }
    \hline
     \textbf{Network Arch.} & \textbf{Classifier} & \textbf{Classifier}\\ 
    \textbf{Neurons/layer} &   \textbf{Accuracy} & \textbf{Accuracy} \\ 
    \textbf{[Layers]} & \textbf{SAT-4 (\%)} & \textbf{SAT-6 (\%)} \\ \hline
    100 [2] & 79.74  & 68.51 \\ \hline
    \textbf{100 [3]} & \textbf{81.78}  & \textbf{76.47} \\ \hline
    100 [4] & 79.802  & 74.44 \\ \hline
    100 [5] & 62.776  & 63.14 \\ \hline
    500 [2] & 68.916 &  60.35 \\ \hline
    500 [3] &  71.674 &  61.12  \\ \hline
    500 [4] & 65.002  & 57.31 \\ \hline
    500 [5] & 64.174  & 55.78 \\ \hline
  \end{tabular}
  \caption{Classification Accuracy of DBN with various architectures on SAT-4 and SAT-6}
  \label{table:DBN_accuracy_SAT_4_and_SAT_6}
\end{table}

\subsection{Convolutional Neural Network}
Convolutional Neural Network (CNN) first introduced in \cite{fukushima:neocognitronbc} is a hierarchical model inspired by the human visual cortical system \cite{Hubel:62}. It was significantly improved and applied to document recognition in \cite{Lecun98gradient-basedlearning}. A committee of 35 convolutional neural nets with elastic distortions and width normalization \cite{Ciresan2012} has produced state-of-the-art results on the MNIST handwritten digits dataset. CNN consists of a hierarchical representation using convolutional layers and fully connected layers, with non-linear transformations and feature pooling. 

They also include local or global pooling layers. Pooling can be implemented in the form of subsampling, averaging, max-pooling or stochastic pooling. Each of these pooling architectures has its own advantages and limitations and numerous studies are in place that investigate the effect of different pooling functions on representation power of the model (\cite{Scherer2010},\cite{ICML2011Saxe_551}). A very important feature of Convolutional Neural Network is weight sharing in the convolutional layers, so that the same filter bank is applied to all pixels in a particular layer; thereby generating sparse networks that can generalize well to unseen data samples while maintaining the representational power inherent in deep hierarchical architectures. 

We investigate the use of different CNN architectures for SAT-4 and SAT-6 as detailed below.

\subsubsection{CNN Results on SAT-4 \& SAT-6}
For CNN, we vary the number of feature maps in each layer as well as the total number of convolutional and subsampling layers. The results from various network configurations with increasing number of maps and layers is enumerated in Table \ref{table:CNN_accuracy_SAT_4_and_SAT_6}. For the experiments, we used both $3{\times}3$ and $5{\times}5$ kernels for the convolutional layers and $3{\times}3$ averaging and max-pooling kernels for the sub-sampling layers. We also use overlapping pooling windows with a stride size of $2$ pixels. The last sub-sampling layer is connected to a fully-connected layer with 64 neurons. The output of the fully-connected layer is fed into a 4-way softmax function that generates a probability distribution over the 4 class labels of SAT-4 and a 6-way softmax for the 6 class labels of SAT-6. In Table \ref{table:CNN_accuracy_SAT_4_and_SAT_6}, the ``Ac-Bs(n)'' notation denotes that the network has a convolutional layer with A feature maps followed by a sub-sampling layer with a kernel of size $B{\times}B$. `n' denotes the type of pooling function in the sub-sampling layer, `a' denotes average pooling while `m' denotes max-pooling. From the table, it can be seen that the smallest networks consistently produce the best results. Also, both for SAT-4 and SAT-6, using networks with convolution kernels of size $3{\times}3$ leads to a significant drop in classifier accuracy. The biggest networks with 50 maps per layer also exhibit significant drop in classifier accuracy.   

\begin{table}[h!]
\centering
\begin{tabular}{ | c | c | c | c |}
    \hline
    \textbf{Network Architecture} & \textbf{Accuracy} & \textbf{Accuracy}\\   
     \textbf{(Convolution kernel size)} & \textbf{SAT-4} & \textbf{SAT-6} \\ 
     &\textbf{(\%)} & \textbf{(\%)}\\ \hline
    \textbf{6c-3s(a)-12c-3s(m) ($5{\times}5$)} & \textbf{86.827} & \textbf{79.063} \\ \hline
    18c-3s(a)-36c-3s(m) ($5{\times}5$) & 82.325 & 78.704\\ \hline
    6c-3s(a)-12c-3s(m)-12c & 81.907  & 76.963\\ 
    -3s(m)($5{\times}5$) & &\\ \hline
    50c-3s(a)-50c-3s(m)-50c & 73.85 & 75.689 \\ 
    -3s(m)($5{\times}5$) & &\\ \hline
    6c-3s(a)-12c-3s(m) ($3{\times}3$) & 73.811  & 54.385 \\ \hline
    6c-3s(m)-12c-3s(m) ($5{\times}5$) & 85.612  & 77.636 \\ \hline
  \end{tabular}
  \caption{Classification Accuracy of CNN with various architectures on SAT-4}
  \label{table:CNN_accuracy_SAT_4_and_SAT_6}
\end{table}     

\subsection{Stacked Denoising Autoencoder}
A Stacked Denoising Autoencoder (SDAE)~\cite{Vincent:2010} consists of a combination of multiple sparse autoencoders, which can be trained in a greedy-layerwise fashion similar to that of Restricted Boltzmann Machines in a DBN. Each autoencoder is associated with a set of weights and biases. In the SDAE, each layer can be trained independent of the other layers. Once trained, the parameters of an autoencoder are frozen in place. The training algorithm is comprised of two phases -- a forward pass phase and a backward pass phase. The forward pass, also called as the encoding phase encodes raw image pixels into an increasingly higher-order representation. The backward pass simply performs the reverse operation by decoding these higher-order features into simpler representations. 
The encoding step is given as:

\begin{equation}
a^{(l)} = f(z^{(l)})
\end{equation}

\begin{equation}
z^{(l+1)}= W^{(l,1)} a^{(l)}+ b^{(l,1)}
\end{equation}

And the decoding step is as follows:

\begin{equation}
a^{(n+l)} = f(z^{(n+l)})
\end{equation}

\begin{equation}
z^{(n+l+1)}= W^{(n-l,2)} a^{(n+l)}+ b^{(n-l,2)}
\end{equation}

The hidden unit activations of the neurons in the deepest layer are used for classification after a supervised fine-tuning using backpropagation.

\subsubsection{SDAE Results on SAT-4 \& SAT-6}
Different network configurations were chosen for the SDAE in a manner similar to that described above for DBN and CNN. The results are enumerated in Table \ref{table:SAE_accuracy_SAT_4_and_SAT_6}. Similar to DBN, each network is trained for a maximum of 500 epochs and the lowest test error is considered for evaluation. As highlighted in the Table, networks with 5 layers and 100 neurons in each layer produce the best results on both SAT-4 and SAT-6. It can be seen from the table that on both datasets, the classifier accuracy initially improves and then drops with increasing number of neurons and layers, similar to that of DBN. Also, the biggest networks with 500 and 2352 neurons in each layer exhibit a significant drop in classifier accuracy.  

\begin{table}[h]
\centering
\begin{tabular}{ | c | c | c | }
    \hline
     \textbf{Network Arch.} & \textbf{Classifier} & \textbf{Classifier}\\ 
    \textbf{Neurons/layer} &   \textbf{Accuracy} & \textbf{Accuracy} \\ 
    \textbf{[Layers]} & \textbf{SAT-4 (\%)} & \textbf{SAT-6 (\%)} \\ \hline
    100 [1] & 75.88  & 74.89 \\ \hline
    100 [2] & 76.854 & 76.12 \\ \hline
    100 [3] & 77.804 & 76.45  \\ \hline
    100 [4] & 78.674 & 76.52  \\ \hline
    \textbf{100 [5]} & \textbf{79.978} & \textbf{78.43} \\ \hline
    100 [6] & 75.766 & 76.72  \\ \hline
    500 [3] & 63.832 & 54.37 \\ \hline
    2352 [2] & 51.766 & 37.121 \\ \hline
  \end{tabular}
  \caption{Classification Accuracy of SDAE with various architectures on SAT-4 and SAT-6}
  \label{table:SAE_accuracy_SAT_4_and_SAT_6}
\end{table}

\section{DeepSat - A Detailed \\
Architectural Overview}
\begin{figure*}
  \centering
    \includegraphics[width=0.8\textwidth]{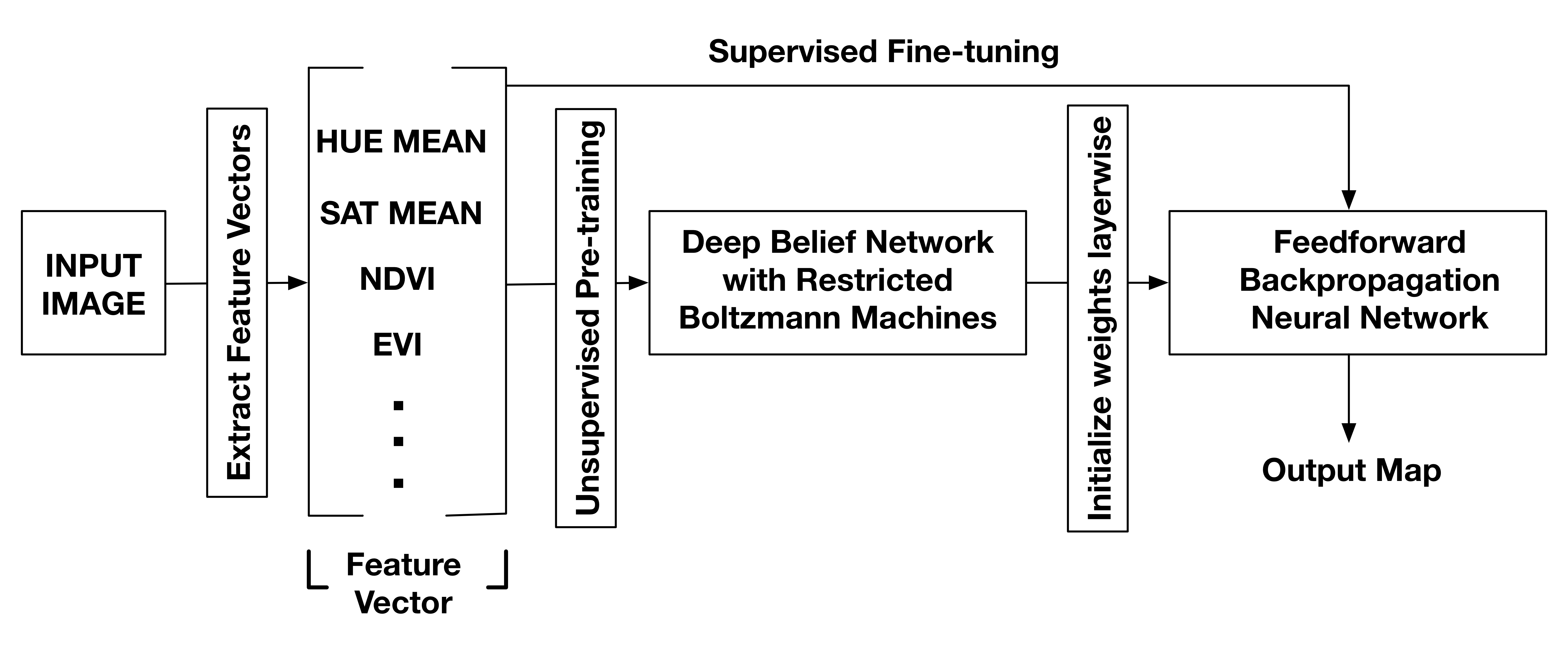}
  \caption{Schematic of the DeepSat classification framework} \label{Fig:1}
\end{figure*}
Figure \ref{Fig:1} schematically describes our proposed classification framework. Instead of the traditional DBN model described in Section \ref{sec:DBN}, which takes as input the multi-channel image pixels reshaped as a linear vector, our classification framework first extracts features from the image which in turn are fed as input to the DBN after normalizing the feature vectors.

\subsection{Feature Extraction}
The feature extraction phase computes 150 features from the input imagery.  The key features that we use for classification are mean, standard deviation, variance, 2nd moment, direct cosine transforms, correlation, co-variance, autocorrelation, energy, entropy, homogeneity, contrast, maximum probability and sum of variance of the hue, saturation, intensity, and NIR channels as well as those of the color co-occurrence matrices. These features were shown to be useful descriptors for classification of satellite imagery in previous studies (\cite{haralick1973}, \cite{Soh99textureanalysis}, \cite{Clausi2002}). Since two of the classes in SAT-4 and SAT-6 are trees and grasslands, we incorporate features that are useful determinants for segregation of vegetated areas from non-vegetated ones. The red band already provides a useful feature for discrimination of vegetated and non-vegetated areas based on chlorophyll reflectance, however, we also use derived features (vegetation indices derived from spectral band combinations) that are more representative of vegetation greenness -- this includes the Enhanced Vegetation Index (EVI \cite{huete2002}), Normalized Difference Vegetation Index (NDVI \cite{rouse1974}, \cite{Tucker1979127}) and Atmospherically Resistant Vegetation Index (ARVI \cite{kaufman1992}).

These indices are expressed as follows:

\begin{equation}
EVI = G\times\frac{NIR-Red}{NIR+c_{red}\times{Red}-c_{blue}\times{Blue}+L}
\end{equation}

Here, the coefficients $G$, $c_{red}$, $c_{blue}$ and $L$ are chosen to be 2.5, 6, 7.5 and 1 following those adopted in the MODIS EVI algorithm \cite{modis}.

\begin{equation}
NDVI = \frac{NIR-Red}{NIR+Red}
\end{equation}

\begin{equation}
ARVI = \frac{NIR-(2\times{Red}-Blue)}{NIR+(2\times{Red}+Blue)}
\end{equation}

The performance of our learner depends to a large extent on the selected features. Some features contribute more than others towards optimal classification. The 150 features extracted are narrowed down to 22 using a feature-ranking algorithm based on Distribution Separability Criterion \cite{Boureau10atheoretical}. Details of the feature ranking method along with the ranking for all the 22 features used in our framework is listed in Section \ref{Section:feature_ranking}. 

\subsection{Data Normalization}
The feature vectors extracted from the training and test datasets are separately normalized to lie in the range $[0,1]$. This is done using the following equation:

\begin{equation}
F_{normalized} = \frac{F-F_{min}}{F_{max}-F_{min}}
\end{equation}

where, $F_{min}$ and $F_{max}$ are computed for a particular feature type over all images in the dataset.

\subsection{Classification}
The set of normalized feature descriptors extracted from the input image is fed into the DBN, which is then trained using \emph{Contrastive divergence} in the same way as explained in Section \ref{sec:DBN}. Once trained the DBN is used to initialize the weights of a feedforward backpropagation neural network. 

The neural network gives an estimate of the posterior probabilities of the class labels, given the input vectors, which is the feature vector in our case. As illustrated in \cite{Bishop1995}, the outputs of a neural network which is obtained by optimizing the sum-squared error-gradient function approximates the average of the class conditional distributions of the target variables

\begin{equation} \label{eq:1}
y_k(x) = \langle t_k|x \rangle = \int t_k p(t_k|x)dt_k
\end{equation}

Here, $t_k$ are the set of target values that represent the class membership of the input vector $x_k$. For a binary classification problem, in order to map the outputs of the neural network to the posterior probabilities of the labeling, we use a single output $y$ and a target coding that sets $t^n=1$ if $x^n$ is from class $C_1$ and $t^n=0$ if $x^n$ is from class $C_2$. The target distribution would then be given as

\begin{equation} \label{eq:2}
p(t_k|x) = \delta{(t-1)}P(C_1|x) + \delta{(t)}P(C_2|x)
\end{equation}

Here, $\delta$ denotes the Dirac delta function which has the properties $\delta(x)=0$ if $x \neq 0$ and

\begin{equation}
\int_{-\infty}^{\infty} \delta(x)\,\mathrm{d}x = 1
\end{equation} 

From \ref{eq:1} and \ref{eq:2}, we get

\begin{equation}
y(x)= P(C_1|x)
\end{equation}        							        

So, the network output $y(x)$ represents the posterior probability of the input vector $x$ having the class membership $C_1$ and the probability of the class membership $C_2$ is given by $P(C_2|x) = 1-y(x)$. This argument can easily be extended to multiple class labels for a generalized multi-class classification problem. 

The feature extraction phase proves to be a useful dimensionality reduction technique that helps improve the discriminative power of the DBN based classifier significantly.

\section{Results and Comparative Studies}
The feature vectors extracted from the dataset are fed into DBNs with different configurations. Since, the feature vectors create a low dimensional representation of the data, so, DeepSat converges to high accuracy even with a much smaller network with fewer layers and very few neurons per layer. This speeds up network training by several orders of magnitude. Various network architectures along with the classification accuracy for DeepSat on the SAT-4 and SAT-6 datasets are listed in Table \ref{table:DeepSat_accuracy_SAT_4_and_6}. For regularization, we again use $L_2$ norm-regularization. From the Table, it is evident that the best performing DeepSat network outperforms the best traditional Deep Learning approach (CNN) by ${\sim}$11\% on the SAT-4 dataset and by ${\sim}$15\% on the SAT-6 dataset.    

We also compare DeepSat with a Random Forest classifier to investigate the advantages gained by unsupervised pre-training in DBN as opposed to the traditional supervised learning in Random Forests. On SAT-4, the Random forest classifier produces an accuracy of 69\% while on SAT-6, it produces an accuracy of 54\%. The highest accuracy was obtained for a forest with 100 trees. Further increase in the number of trees did not yield any significant improvement in classifier accuracy. It can be easily seen that the various Deep architectures produce better classification accuracy than the Random Forest classifier which relies solely on supervised learning.    

\begin{table}[h!]
\centering
\begin{tabular}{ | c | c | c | }
    \hline
    \textbf{Network Arch.} & \textbf{Classifier} & \textbf{Classifier}\\ 
    \textbf{Neurons/layer} &   \textbf{Accuracy} & \textbf{Accuracy} \\ 
    \textbf{[Layers]} & \textbf{SAT-4 (\%)} & \textbf{SAT-6 (\%)} \\ \hline
    10 [2] & 96.585 & 91.91\\ \hline
    10 [3] & 96.8  & 87.716 \\ \hline
    20 [2] & 97.115 & 86.21 \\ \hline
    20 [3] & 95.473 & 93.42 \\ \hline
    \textbf{50 [2]} & \textbf{97.946} &\textbf{93.916} \\ \hline
    50 [3] & 97.654 & 92.65 \\ \hline
    100 [2] & 97.292 & 89.08 \\ \hline
    100 [3] & 95.609 & 91.057 \\ \hline
  \end{tabular}
  \caption{Classification Accuracy of DeepSat with various network architectures on SAT-4 and SAT-6}
  \label{table:DeepSat_accuracy_SAT_4_and_6}
\end{table}

\begin{figure*}
\centering
\subfigure[Distribution of NIR on the SAT-4 classes]{\includegraphics[width=0.45\linewidth]{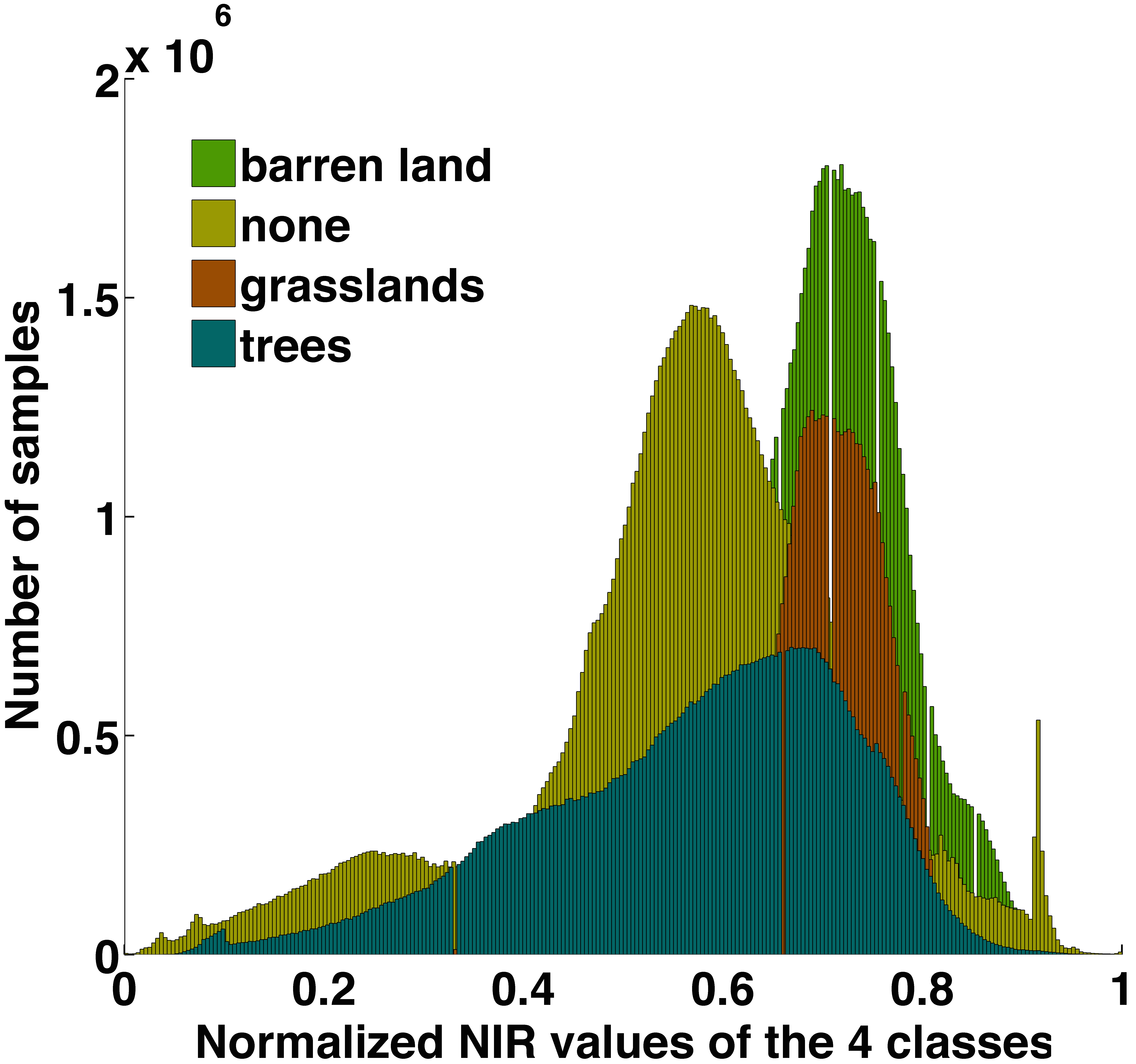}}
\subfigure[Distribution of a sample DeepSat feature (Autocorrelation of Hue Color co-occurance matrix) on the SAT-4 classes]{ \includegraphics[width=0.45\linewidth]{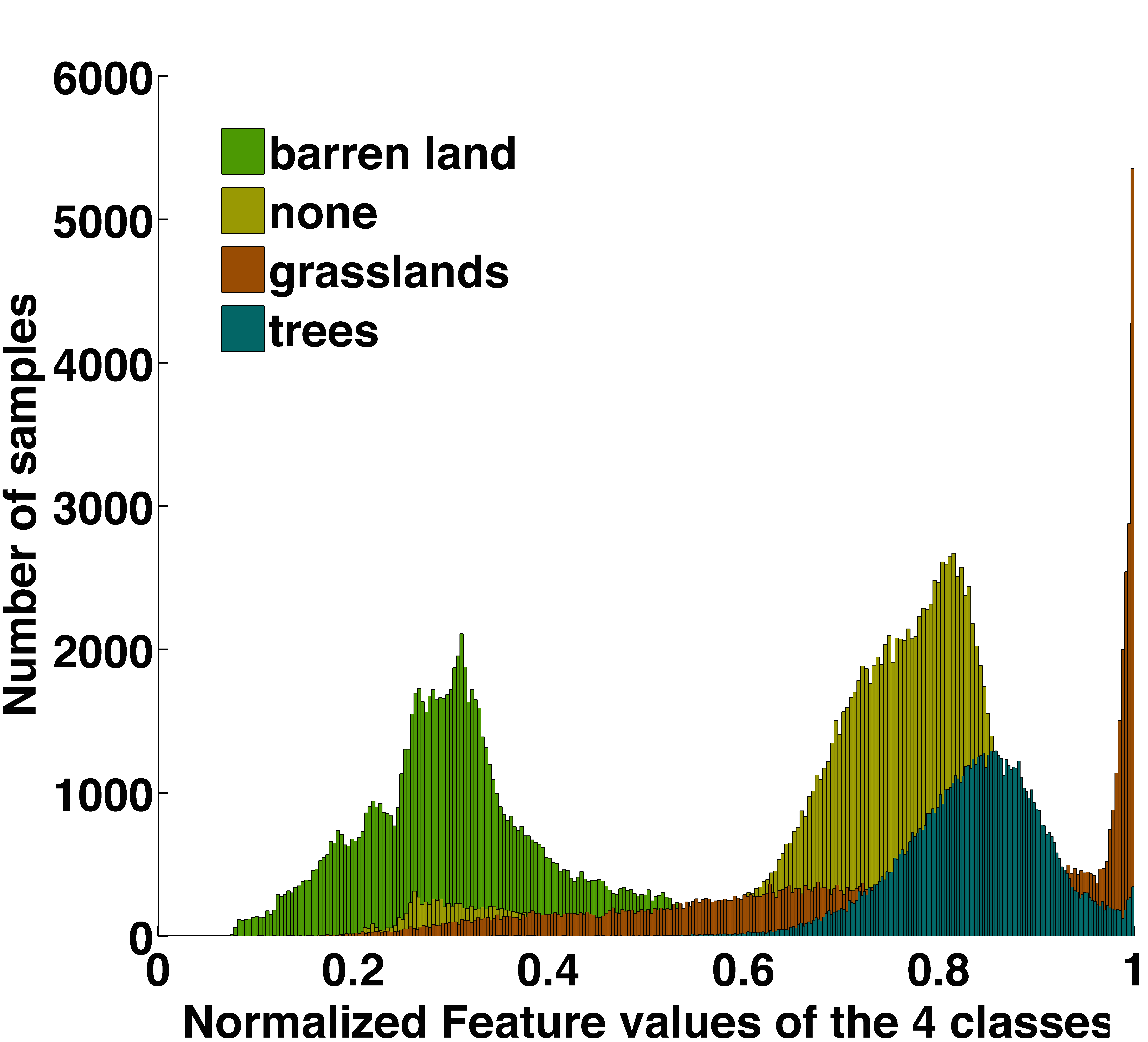}}
\caption{Distributions of the raw NIR values for traditional Deep Learning Algorithms and a sample DeepSat feature for various classes on SAT-4 (\emph{Best viewed in color})}
\label{fig:distributions}
\end{figure*}

\section{Why Traditional Deep Architectures are not enough for SAT-4  \& SAT-6?}
While traditional Deep Learning approaches have produced state-of-the-art results for various pattern recognition problems like handwritten digit recognition \cite{mnist}, object recognition \cite{Krizhevsky09learningmultiple}, face recognition \cite{DeepFace2013}, etc., but satellite datasets have high intra and inter-class variability and the amount of labeled data is much smaller as compared to the total size of the dataset. Also, higher-order texture features are a very important discriminative parameter for various landcover classes. On the contrary, shape/edge based features which are predominantly learned by various Deep architectures are not very useful in learning data representations for satellite imagery. This explains the fact why traditional Deep architectures are not able to converge to the global optima even for reasonably large as well as Deep architectures. 

Also, spatially contextual information is another important parameter for modeling satellite imagery. In traditional Deep Learning approaches like DBN and SDAE, the relative spatial information of the pixels is lost. As a result the orderless pool of pixel values which acts as input to the Deep Networks lack sufficient discriminative power to be well-represented even by very big and/or deep networks. CNN however, involves feature-pooling from a local spatial neighborhood, which justifies its improved performance over the other two algorithms on both SAT-4 and SAT-6. Even though our approach extracts an orderless pool of feature vectors, the spatial context is already well-represented in the individual feature values themselves. We substantiate our arguments about the effectiveness of our feature extraction approach from a statistical point of view as detailed in the analysis below.  
\begin{table}[h]
\centering
\begin{tabular}{ | c | c | c | c | }
    \hline
 &   & \textbf{Dist. b/w} & \textbf{Standard} \\ 
 &   & \textbf{Means} &   \textbf{Deviations}  \\ \hline
 \multirow{2}{*}{\raisebox{-.15in}{\rotatebox{90}{\tiny{SAT-4}}}} & Raw Images & 0.1994 & 0.1166 \\ 
  & DeepSat Features & \textbf{0.8454} & \textbf{0.0435} \\ \hline
 \multirow{2}{*}{\raisebox{-.15in}{\rotatebox{90}{\tiny{SAT-6}}}} & Raw Images & 0.3247 & 0.1273 \\
  & DeepSat Features & \textbf{0.9726} & \textbf{0.0491} \\ \hline
  \end{tabular}
  \caption{Distance between Means and Standard Deviations for raw image values and DeepSat feature vectors for SAT-4 and SAT-6}
  \label{table:Distribution_mean_and_sd}
\end{table}

\subsection{A Statistical Perspective based on Distribution Separability Criterion}\label{statistical_perspective}
Improving classification accuracy can be viewed as maximizing the separability between the class-conditional distributions. Following the analysis presented in \cite{Boureau10atheoretical}, we can view the problem of maximizing distribution separability as maximizing the distance between distribution means and minimizing their standard deviations. Figure \ref{fig:distributions} shows the histograms that represent the class-conditional distributions of the NIR channel and a sample feature extracted in the DeepSat framework. As illustrated in Table \ref{table:Distribution_mean_and_sd}, the features extracted in DeepSat have a higher distance between means and a lower standard deviation as compared to the original image distributions, thereby ensuring better class separability.

\subsubsection{Feature Ranking}\label{Section:feature_ranking}
Following the analysis proposed in Section \ref{statistical_perspective} above, we can derive a metric for the Distribution Separability Criterion as follows:
\begin{equation}
D_s = \frac{\overbar{\lVert \delta_{mean} \rVert }}{\overbar{\delta_{\sigma}}}
\end{equation}
where $\overbar{\lVert \delta_{mean} \rVert }$ indicates the mean of distance between means and $\overbar{\delta_{\sigma}}$ indicates the mean of standard deviations of the class conditional distributions. Maximizing $D_s$ over the feature space, a feature ranking can be obtained. Table \ref{table:Feature_ranking} shows the ranking of the various features used in our framework along with the values of the corresponding distance between means $\overbar{\lVert \delta_{mean} \rVert }$, standard deviation  $\overbar{\delta_{\sigma}}$ and Distribution Separability Criterion $D_s$.
\begin{table}[h!]
\centering
\begin{tabular}{ |c|c|c|c|c|}
    \hline
\textbf{Rank} & \textbf{Feature} & $\boldsymbol{\overbar{\lVert \delta_{mean} \rVert }}$  &  $\boldsymbol{\overbar{\delta_{\sigma}}}$ & $\boldsymbol{D_s}$  \\[1.5ex] \hline
 1&I CCM mean & 0.4031  & 0.1371  &  2.9403 \\ \hline
 2 & H CCM sosvh  &  0.2359   & 0.0928  &  2.5413   \\ \hline
 3 & H CCM autoc   &  0.2334    & 0.1090  &  2.1417  \\ \hline
 4 & S CCM mean    &   0.0952   & 0.0675  & 1.4099   \\ \hline
 5 &  H CCM mean    &  0.0629   & 0.0560  &  1.1237   \\ \hline
 6 &   SR    &  0.0403   & 0.0428  &   0.9424  \\ \hline
 7 &   S CCM     &   0.0260   & 0.0312  &  0.8354  \\ 
   &   2nd moment     &      &   &   \\ \hline
  8 &   I CCM   &  0.0260    & 0.0312  & 0.8354   \\
    &   2nd moment     &      &   &   \\ \hline
   9  &   I 2nd moment    &  0.0260    & 0.0312  &   0.8345 \\ \hline
    10   &  I variance    &   0.0260   & 0.0312  &  0.8345  \\ \hline
    11   &   NIR std    &  0.0251    & 0.0315  &  0.7980  \\ \hline
   12    &   I std     &  0.0251    & 0.0314  &  0.7968  \\ \hline
    13    &  H std      &  0.0252    & 0.0317  &  0.7956  \\ \hline
    14    &   H mean      &  0.0240   & 0.0314  &  0.7632   \\ \hline
     15   &    I mean      &  0.0254   &  0.0336 &  0.7541  \\ \hline
     16   &    S mean       &   0.0232   & 0.0319  &  0.7268  \\ \hline
      17     &   I CCM       &   0.0378    &  0.0522 &  0.7228  \\ 
        &   covariance     &      &   &   \\ \hline
       18     &   NIR mean      &  0.0246     & 0.0351  &  0.6997 \\ \hline
       19      &   ARVI      &  0.0229    &  0.0345 &   0.6622 \\ \hline
        20      &   NDVI      &   0.0215    & 0.0326  &  0.6594 \\ \hline
         21     &    DCT      &   0.0344    &  0.0594 &  0.5792 \\ \hline
          22     &    EVI      &    0.0144   & 0.0450  & 0.3207  \\ \hline
  \end{tabular}
  \caption{Ranking of features based on Distribution Separability Criterion for SAT-6}
  \label{table:Feature_ranking}
\end{table}

\subsubsection{Distribution Separability and Classifier Accuracy}
\begin{figure*}
\centering
\subfigure[Distribution Separability Criterion of DBN]{\includegraphics[width=0.45\linewidth]{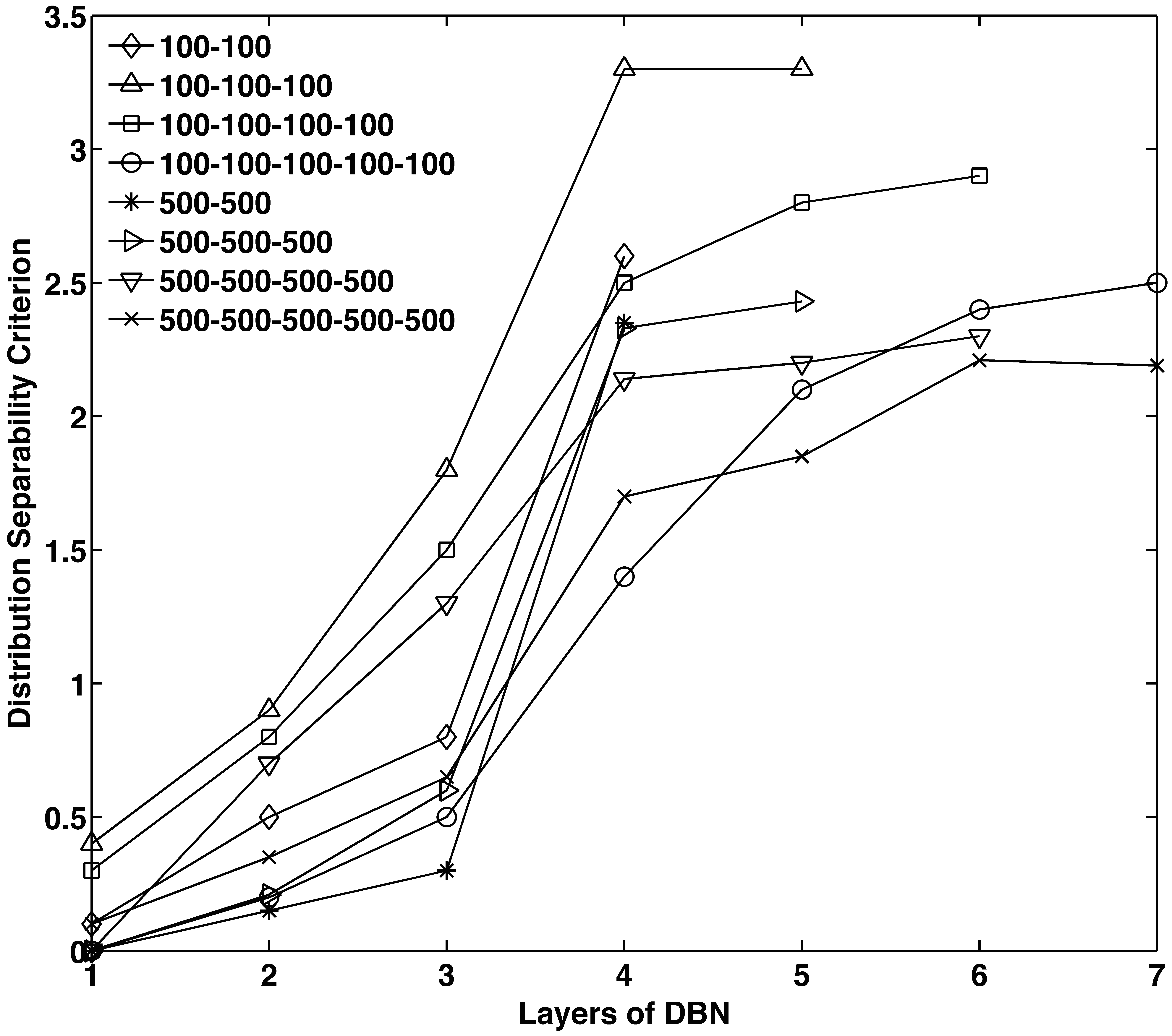}}
\subfigure[Distribution Separability Criterion of DeepSat]{ \includegraphics[width=0.45\linewidth]{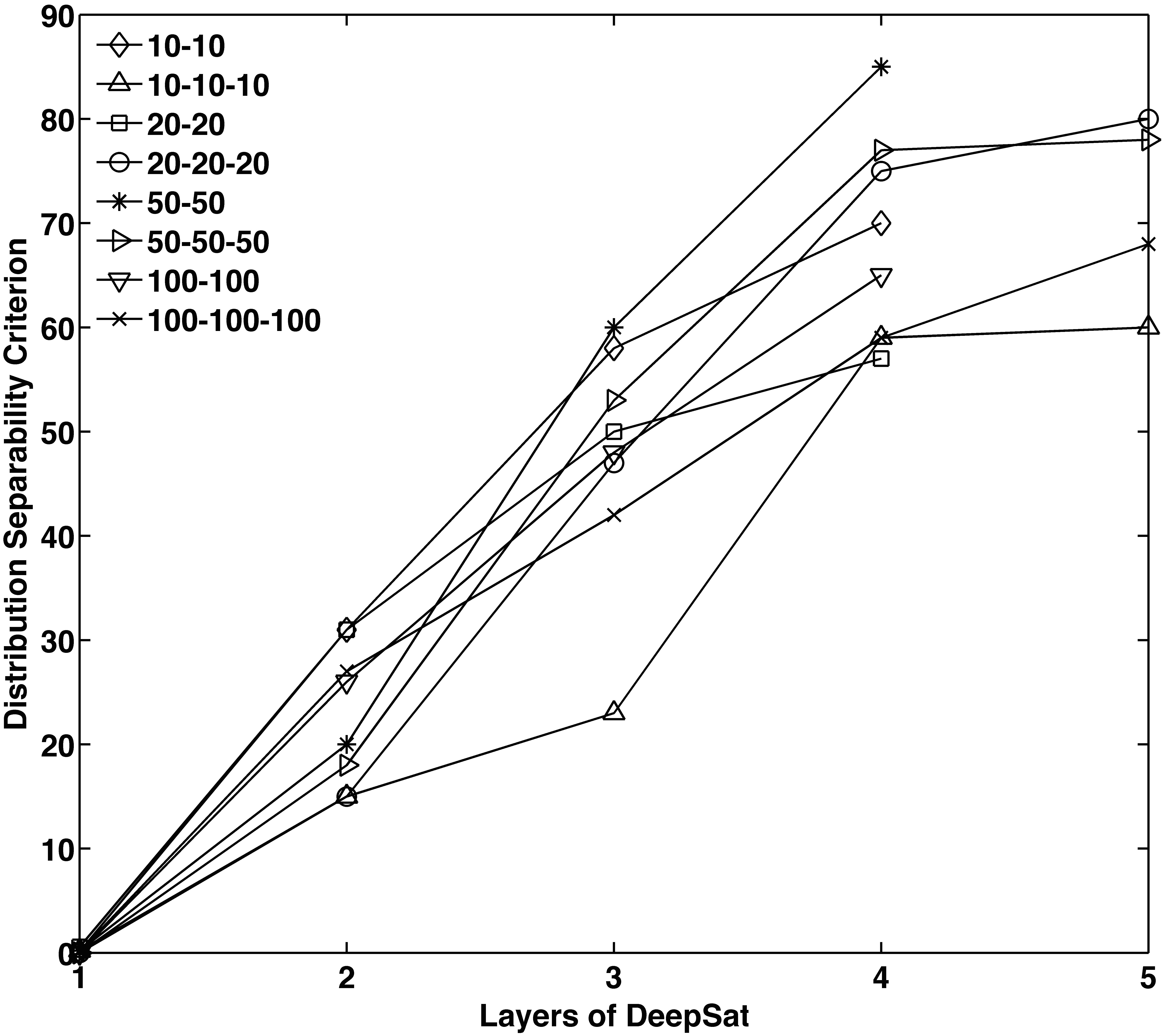}}
\caption{Distribution Separability Criterion of the neurons in the layers of a DBN and DeepSat with various architectures on SAT-6}
\label{dist_sep_neurons}
\end{figure*}

In order to analyze the improvements achieved in the learning framework due to the feature extraction step, we measured the Distribution Separability of the mean activation of the neurons in each layer of the DBN and that of DeepSat. The results are noted in Figure \ref{dist_sep_neurons}. It can be seen that the mean activation learned by each layer of DeepSat exhibit a significantly higher distribution separability (by several orders of magnitude) than the neurons of a DBN. This justifies the significant improvement in performance of DeepSat (using the features) as compared to the DBN based framework (using the raw pixel values as input). Also, a comparison of Figure \ref{dist_sep_neurons} with Table \ref{table:DBN_accuracy_SAT_4_and_SAT_6} and Table \ref{table:DeepSat_accuracy_SAT_4_and_6} shows that the distribution separabilities using the various architectures of the DBN and DeepSat are positively correlated to the final classifier accuracy. This justifies the effectiveness of our distribution separability metric $D_s$ as a measure of the final classifier accuracy.

\section{What is the difference between MNIST, CIFAR-10 and SAT-6 in terms of dimensionality?}  
We argue that handwritten digit datasets like MNIST and object recognition datasets like CIFAR-10 lie on a much lower dimensional manifold than the airborne SAT-6 dataset. Hence, even if Deep Neural Networks can effectively classify the raw feature space of object recognition datasets but the dimensionality of the airborne image datasets is such that Deep Neural Networks cannot classify them. In order to estimate the dimensionality of the datasets, we use the concept of \emph{intrinsic dimension}\cite{Ceruti20142569}.   

\subsection{Intrinsic Dimension Estimation using the DanCo algorithm}
To estimate the intrinsic dimension of a dataset, we use the DANCo algorithm \cite{Ceruti20142569}. It uses the complementary information provided by the normalized nearest neighbor distances and angles calculated on pairs of neighboring points.

Taking 10 rounds of 1000 random samples and averaging, we obtain the intrinsic dimension for the MNIST, CIFAR-10 and SAT-6 datasets and the Haralick features extracted from the SAT-6 dataset. The results are listed in Table \ref{table:Intrinsic_Dimension}.

\begin{table}[h]
\centering
\begin{tabular}{ | c | c | c | }
    \hline
 \textbf{Dataset}  & \textbf{Intrinsic Dimension} \\  \hline
 MNIST & 16  \\  \hline
 CIFAR-10 & 17 \\ \hline
SAT-6 & 115 \\ \hline
Haralick Features extracted from SAT-6 & \textbf{4.2} \\ \hline
  \end{tabular}
  \caption{Intrinsic Dimension estimation using DANCo on the MNIST, CIFAR-10, and SAT-6 datasets and the Haralick features extracted from the SAT-6 dataset.}
  \label{table:Intrinsic_Dimension}
\end{table}

So, it can be seen that the intrinsic dimensionality of the SAT-6 dataset is orders of magnitude higher than that of MNIST. So, a deep neural network finds it difficult to classify the SAT-6 dataset because of its intrinsically high dimensionality. However, as seen in the equation above, the features extracted from SAT-6 have a much lower intrinsic dimensionality and lie on a much lower dimensional manifold than the raw vectors and hence can be classified even by networks with relatively smaller architectures.

\subsection{Visualizing Data in an n-dimensional space}

We can visualize the data as distributed in an n-dimensional unit hypersphere \\
Volume of the sphere, 
\begin{equation}
{V}_{sphere} = \frac{\pi^{\frac{n}{2}}}{\Gamma(\frac{n}{2}+1)}R^{n} = \frac{\pi^{\frac{n}{2}}}{\Gamma(\frac{n}{2}+1)}
\end{equation}
for n-dimensional Euclidean space and $\Gamma$ is Euler's gamma function. Now, the total volume of the n-dimensional space can be accounted by the volume of an n-dimensional hypercube of length 2 embedding the hypersphere, i.e, Volume of the n-cube, 
\begin{equation}
{V}_{cube} = {R}^{n} = 2^n
\end{equation} 
So, the relative fraction of the data points which lie on the sphere as compared to the data points on the n-dimensional embedding space is given as
\begin{equation}
{V}_{relative} = \frac{{V}_{sphere}}{{V}_{cube}} = \frac{\pi^{\frac{n}{2}}}{{2}^{n}\Gamma(\frac{n}{2}+1)}
\end{equation}
\begin{equation}
{V}_{relative} \to 0 ~\text{as}~ n \to \infty
\end{equation} 
This means that as the dimensionality of sample data approaches $\infty$, the spread or scatter of the data points approaches 0 with respect to the total search space. As a result, various classification and clustering algorithms lose their discriminative power in higher dimensional feature spaces.   

\section{Related Work}
Present classification algorithms used for Moderate-resolution Imaging Spectroradiometer (MODIS)(500-m) \cite{Friedl_MCD_Coll5Validation2009} or Landsat(30-m) based land cover maps like NLCD \cite{wickham2013} produce accuracies of 75\% and 78\% resp. The relatively lower resolution of the datasets makes it difficult to analyze the performance of these algorithms for 1-m imagery. A method based on object detection using Bayes framework and subsequent clustering of the objects using Latent Dirichlet Allocation was proposed in \cite{Vaduva2012}. However, their approach detects object groups at a higher level of abstraction like parking lots. Detecting the objects like cars or trees in itself is not addressed in their work. A deep convolutional hierarchical framework was proposed recently by \cite{romerounsupervised}. However, they report results on the AVIRIS Indiana's Indian Pines test site. The spatial resolution of the dataset is limited to 20m and it is difficult to evaluate the performance of their algorithm for object recognition tasks at a higher resolution. An evaluation of various feature learning strategies was done in \cite{Tokarczyk2012}. They evaluated both feature extraction techniques as well as classifiers like DBN and Random Forest for various aerial datasets. However, since the training data was significantly limited, the DBN was not able to produce any improvements over Random Forest even when raw pixel values were fed into the classifier. In contrast, our study shows that DBNs can be better classifiers when there is significant amount of training data to initialize the neural network at a global error basin.  

%

\section{Conclusions and Future Directions}

Our semi-supervised learning framework produces an accuracy of 97.95\% and 93.9\% on the SAT-4 and SAT-6 datasets and significantly outperforms the state-of-the-art by ${\sim}$11\% and ${\sim}$15\% respectively. The Feature extraction phase is inspired by the remote sensing literature and significantly improves the discriminative power of the framework. For satellite datasets, with inherently high variability, traditional deep learning approaches are unable to converge to a global optima even with significantly big and deep architectures. A statistical analysis based on Distribution Separability Criterion justifies the effectiveness of our feature extraction approach.    

We plan to investigate the use of various pooling techniques like SPM \cite{Lazebnik:2006} as well as certain sparse representations like sparse coding \cite{Lee07efficientsparse} and Hierarchical representations like Convolutional DBN \cite{Lee:2009:CDBN} to handle satellite datasets. We believe that SAT-4 and SAT-6 will enable researchers to learn better representations for satellite datasets and create benchmarks for the classification of satellite imagery.

%
%


\section{Acknowledgments}
The project is supported by NASA Carbon Monitoring System through Grant \#NN-\\H14ZDA001-N-CMS and Army Research Office (ARO) under Grant \#W911NF1-\\010495. We are grateful to the United States Department of Agriculture for providing us the National Agriculture Imagery Program (NAIP) airborne imagery dataset for the Continental United States.

This research was partially supported by the Cooperative Agreement Number NASA-NNX12AD05A, CFDA Number 43.001, for the project identified as "Ames Research Center Cooperative for Research in Earth Science and Technology (ARC-CREST)". Any opinions findings, and conclusions or recommendations expressed in this material are those of the authors and do not necessarily reflect that of NASA, ARO or the United States Government. 


%

\bibliographystyle{abbrv}
\bibliography{References}
%
%

\end{document}